\documentclass[11pt]{article}

\usepackage[preprint]{acl}
\usepackage{times}
\usepackage{latexsym}
\usepackage[T1]{fontenc}
\usepackage[utf8]{inputenc}
\usepackage{microtype}
\usepackage{booktabs}
\usepackage{amsmath}
\usepackage{graphicx}
\usepackage{xurl}
\usepackage{placeins}

\title{From Agent Failures to Text Policies:\\
What Works and What Breaks}
\author{Jaideep Ray \\
Independent Researcher \\
\texttt{jaray@acm.org}
\And
Ankit Goyal \\
Independent Researcher \\
\texttt{ankit@goyalankit.com}}

\begin{document}
\maketitle

\begin{abstract}
TextGrad improves language-model systems by revising text from feedback. Its core thesis is that natural-language feedback can act as a gradient for optimizing text components without changing model weights. Applying it to agents is harder because feedback arrives only after a sequence of actions, making it difficult to identify which decision caused failure. We study this problem by separating the ability to follow a useful policy from the ability to learn that policy from experience. Our main finding is a clear gap between these two abilities. Human-written policies improve two frozen 7B agents on TextWorldExpress by 5.0 success points, showing that useful policy text exists. However, policies generated from agent trajectories do not reliably outperform fixed prompting, even with richer traces, counterfactual evidence, or iterative GEPA search. The main challenge for agent-level TextGrad is therefore not executing textual policy updates, but reliably generating and selecting them from experience.
\end{abstract}

\section{Introduction}

Many deployed language agents cannot be fine-tuned.
Their weights may be frozen, available only through an API, or too expensive to update.
Their instructions, however, remain editable.
This makes text optimization an appealing way to improve an agent without changing its model.

TextGrad follows this idea.
It treats text as an optimizable component and revises that text from natural-language feedback \citep{yuksekgonul2025textgrad}.
The idea is straightforward for an input-output system because the output can be judged directly.
It is harder for an agent.
An agent takes a sequence of actions, and an early mistake changes every state that follows.
A final failure therefore does not reveal which action should change or what lesson will transfer to another task.

We study this problem with frozen language agents and short textual policies.
After a training task, a learner reads the agent's trajectory and proposes a reusable rule.
The rule is then frozen and tested on new tasks.
We call this prototype RulePI.
Our goal is not to present another prompt optimizer.
It is to identify where this kind of optimization succeeds and where it breaks.

We separate the process into four questions:
\begin{itemize}
    \setlength{\itemsep}{0pt}
    \setlength{\parsep}{0pt}
    \setlength{\topsep}{2pt}
    \setlength{\partopsep}{0pt}
    \item \textbf{Execution:} can the frozen agent follow a useful rule?
    \item \textbf{Evidence:} does the trajectory show which decision mattered?
    \item \textbf{Update:} can the learner turn that evidence into a reusable rule?
    \item \textbf{Selection:} can validation keep helpful rules and reject harmful ones?
\end{itemize}
An end-to-end learner fails if any one answer is no.

Our main finding is a gap between execution and learning.
Human-written policies improve two real 7B agents on TextWorldExpress by 5.0 success points.
The agents can therefore use good policy text.
Yet rules learned from their trajectories do not reliably beat fixed prompting.
This remains true when we provide richer step-by-step traces, compare actions from the same state, or use official GEPA search.
The evidence points to a specific bottleneck: the tested systems do not reliably turn experience into a good rule and decide whether to keep it.

\section{Related Work}

Several methods improve agents by storing language rather than changing model weights.
Reflexion writes feedback for another attempt on the same task \citep{shinn2023reflexion}.
ExpeL extracts lessons from training experience and retrieves them later \citep{zhao2023expel}.
RulePI asks a narrower question: can a lesson from past failures become a small policy that helps on new tasks without another test-time attempt?

TextGrad, ProTeGi, and PACE revise prompts from textual critiques \citep{yuksekgonul2025textgrad,pryzant2023protegi,dong2024pace}.
GEPA extends this idea with repeated mutation and Pareto selection, while RAPOA and Feedback Descent use richer analysis or pairwise feedback \citep{agrawal2026gepa,fernandes2026rapoa,lee2025feedbackdescent}.
Kintsugi also edits an agent's control logic \citep{cao2026kintsugi}.
We directly test official GEPA because it is a strong frozen-weight optimizer.

Other methods update model weights.
StarPO, multi-module GRPO, JERP, and experiential RL learn from trajectories through training \citep{wang2025ragen,ziems2026mmgrpo,ye2026jerp,shi2026erl}.
HiPER and HCAPO assign credit to intermediate planning or execution steps before updating weights \citep{peng2026hiper,tan2026hcapo}.
LangMARL produces language-based credit for multiple agents \citep{yao2026langmarl}.
These methods offer stronger ways to locate important decisions, but they answer a different question because they change weights or study multi-agent learning.

\section{Agent-Level TextGrad}
\label{sec:method}

Let $x$ be the current text policy and let $\theta$ be the frozen model weights.
One update has three steps:
\begin{align}
    \tau &\sim \pi_\theta(\cdot\mid x), \\
    g_{\mathrm{text}} &= D(\tau,R(\tau)), \\
    x' &= U(x,g_{\mathrm{text}}),
\end{align}
The agent first produces a trajectory $\tau$ under policy $x$.
The diagnosis function $D$ turns the trajectory and its reward into textual feedback $g_{\mathrm{text}}$.
The update function $U$ uses that feedback to propose a new policy $x'$.
We freeze $x'$ before test and reuse it across held-out tasks.
Unlike a retry controller, the policy sees no failure from the current test task before acting.

\begin{figure*}[t]
\centering
\includegraphics[width=0.98\textwidth]{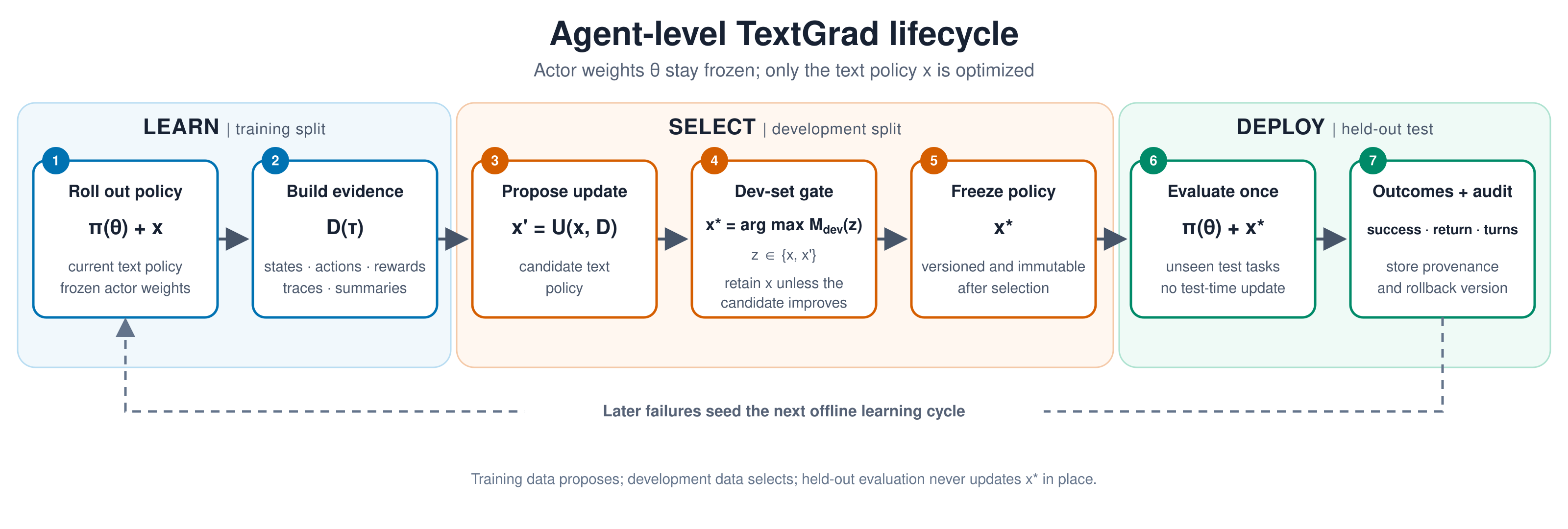}
\caption{Agent-level TextGrad lifecycle. Actor weights remain frozen while training evidence proposes text-policy updates. Development tasks select and freeze the policy, which is evaluated once on unseen held-out tasks without test-time updates.}
\label{fig:lifecycle}
\end{figure*}

We use four tests to locate failure.
First, a \textbf{capacity test} supplies human-written rules with the correct benchmark-family label.
This is privileged information.
It asks whether the actor can use a good rule, not whether the learner can discover one.
Second, an \textbf{evidence test} compares short failure summaries with context-bounded traces that align each observation, action, and reward.
Third, a \textbf{counterfactual test} compares two actions taken from the same failing state.
This removes many differences that confound two complete retries.
Finally, a \textbf{search test} runs official GEPA, which proposes and selects policies over several rounds.
Figure~\ref{fig:lifecycle} shows how these pieces fit into one lifecycle while keeping training, development, and held-out deployment separate.

The single-rule learner may produce one typed rule of at most 80 words for each model and task family.
The eight human-written TextWorldExpress rules contain 15--38 words each (mean 26.4; 211 words in total).
Thus, each individual target strategy fits within the learner's limit, although the capacity condition receives the full labeled library.
This makes every update easy to inspect, but it also limits what the learner can express.
The appendix gives the exact evidence limits, prompts, and selection rules.

\section{Experimental Setup}

\paragraph{Agents and environments.}
We test Qwen2.5-7B-Instruct and Mistral-7B-Instruct-v0.3 with deterministic decoding.
The environments are TextWorld, TextWorldExpress, and TextArena \citep{cote2018textworld,jansen2022textworldexpress,guertler2025textarena}.
Our first study contains 6,160 test records across the two models, three benchmarks, and seven conditions.
The clearest real-agent capacity comparison comes from TextWorldExpress, so the main text centers on that benchmark.
The full cross-benchmark results appear in the appendix.

\paragraph{Policies and baselines.}
The fixed baseline uses the original instructions.
The human-policy condition adds rules written for each TextWorldExpress family.
Because both the rules and their family routing are supplied by people, this condition is a capacity test rather than an autonomous learning method.
The diagnostic-retry control gives a failed task one more attempt with the matching human rule.
The learned conditions instead generate rules from summaries, step traces, matched counterfactual branches, or GEPA search.
All policies use separate training, development, and test seeds and are frozen before test.

\paragraph{Evaluation.}
We pair methods on the same environment, seed, and target side.
We report 95\% cluster-bootstrap intervals from 10,000 resamples.
The counterfactual study evaluates 320 paired episodes per policy and gives both learners 102 extra rollouts.
The GEPA study evaluates 160 paired episodes per method after a nominal 320 search calls per model.
These intervals are exploratory because the number of model-family clusters is small.
Our primary capacity estimate pools both models and all eight TextWorldExpress families.
Per-model rows are consistency checks.
Per-family estimates are reported separately in the appendix with unadjusted hierarchical intervals and are not used for confirmatory claims.

\begin{table}[t]
\centering
\scriptsize
\setlength{\tabcolsep}{4pt}
\begin{tabular}{lrrr}
\toprule
Actor & Fixed & Human policy & $\Delta$ [95\% CI] \\
\midrule
Qwen-7B & 17.50 & 21.25 & +3.75 [1.25, 7.50] \\
Mistral-7B & 13.75 & 20.00 & +6.25 [1.25, 13.75] \\
Pooled & 15.63 & 20.63 & +5.00 [1.88, 9.38] \\
\bottomrule
\end{tabular}
\caption{TextWorldExpress success (\%). The two actor rows are per-model estimates; the pooled row is the primary capacity estimate across two models and eight families. These are not per-family effects. The rules and family routing are privileged.}
\label{tab:capacity}
\end{table}

\section{Results}

\subsection{Frozen Agents Can Use Good Policy Text}

Table~\ref{tab:capacity} answers the first question.
Human-written rules raise pooled TextWorldExpress success from 15.63\% to 20.63\%, a gain of 5.0 points (CI: +1.88 to +9.38).
They also reduce interaction by 2.64 turns on average (CI: $-5.15$ to $-0.41$).
Both real 7B models improve.
The pooled row is the estimate used in our claim.
The model rows show that the direction is not driven by one actor; they do not establish uniform gains across task families.

This result is intentionally limited.
People wrote the rules and supplied the correct task-family routing.
The experiment does not show that the system learned those rules.
It shows something more basic but necessary: useful text policies exist, and the frozen agents can follow them.

\subsection{Trajectory-Based Learning Does Not Recover the Gain}

We next ask whether the system can learn similarly useful rules from experience.
Table~\ref{tab:learning-diagnosis} summarizes three increasingly strong attempts.

\begin{table}[t]
\centering
\scriptsize
\setlength{\tabcolsep}{2.0pt}
\begin{tabular}{p{0.20\columnwidth}p{0.30\columnwidth}p{0.40\columnwidth}}
\toprule
Test & Comparison & Success $\Delta$ \\
\midrule
Capacity & Human $-$ fixed & +5.00 [1.88, 9.38] \\
Evidence & Trace $-$ summary & +3.75 [0.00, 8.12] \\
End to end & Trace $-$ fixed & $-1.88$ [$-5.62$, 1.88] \\
Credit & Branch $-$ retry & +0.31 [0.00, 0.94] \\
End to end & Branch $-$ fixed & $-1.25$ [$-3.75$, 0.00] \\
Search & GEPA $-$ fixed & +1.25 [$-3.13$, 5.63] \\
\bottomrule
\end{tabular}
\caption{Pooled TextWorldExpress results in percentage points with paired 95\% intervals. Each row pools two models and eight families within its own matched study. Rows are neither per-family effects nor one leaderboard.}
\label{tab:learning-diagnosis}
\end{table}

First, step-by-step traces help more than compact summaries.
They improve success by 3.75 points over summaries.
However, the trace policy is still 1.88 points below fixed prompting.
The learner benefits from richer evidence but does not turn that evidence into a better policy.

Second, same-prefix branches give a cleaner comparison than full retries.
They compare two actions from the same state while spending the same 102 extra rollouts as the retry control.
The resulting policy is 0.31 points above retry learning, but 1.25 points below fixed prompting.
This is weak directional evidence for better credit, not an end-to-end improvement.

Third, multi-round search does not reliably close the gap at the tested budget.
Official GEPA reaches 20.00\% success versus 18.75\% for fixed prompting after 642 construction rollouts and 62 reflection calls.
The gain is 1.25 points, but its interval spans zero.
The model-level effects also disagree: Qwen falls by 1.25 points while Mistral rises by 3.75 points.

Across all three tests, better evidence or broader search changes the policy, but does not produce a reliable held-out gain.
A manual audit of the 32 directly comparable summary and trace rules helps explain why: only seven are grounded, executable strategies; nine are plausible but too generic; and sixteen are shortcuts, semantically wrong, or malformed.
The full taxonomy and examples appear in the appendix.

\subsection{Two Transitions Limit End-to-End Learning}

The counterfactual study narrows the diagnosis.
A matched branch can show that one action is preferable at a particular state, but a deployable policy must also say when that preference should apply again.
This requires the learner to recover a trigger, an action, a scope, and an exception from a local comparison.
The proposer often loses that structure: it copies an instance detail, broadens a local contrast into unsupported advice, or produces a rule too vague to change behavior.
Cleaner credit assignment therefore does not automatically produce a reusable policy update.

Selection is a separate transition.
A candidate may improve the situations that motivated it while degrading other tasks, and a pooled development score gives limited evidence about that trade-off.
A reliable gate must evaluate support, consistency, and regressions across the relevant task families, not simply choose the candidate with the best observed mean.
Selecting the best available candidate is not the same as establishing that it improves on the current policy.

The pipeline therefore has two distinct interfaces: trajectory evidence must become a properly scoped policy, and that policy must earn deployment through calibrated validation.
The capacity test shows that the actor can execute a concise policy once these decisions are supplied.
The unresolved challenge is making both decisions reliably from experience.

\section{Discussion}

The experiments separate policy capacity from policy induction.
The capacity test shows that frozen agents can improve when given a short, family-specific policy.
The learned-policy studies do not establish a comparable held-out improvement.
Because the human rules are only 15--38 words each, the 80-word limit alone does not explain this gap.
The evidence instead points to two linked difficulties: deriving a correct reusable update from a trajectory and deciding whether to retain it from limited development data.

Richer evidence improves intermediate comparisons.
Step-aligned traces outperform compact summaries, and same-prefix branches isolate action differences more directly than full retries.
Yet neither intervention consistently increases held-out completion after rule generation and selection.
GEPA expands the update space and uses iterative search, but its pooled held-out interval includes zero.
These results distinguish evidence quality from end-to-end policy improvement: better evidence can clarify the update target, but its value depends on the generation and selection procedures that follow.

The results motivate two priorities.
Update operators should preserve the comparison supplied by matched branches and generate multiple typed candidates or compositional policy sets.
Selectors should use paired, family-aware validation, uncertainty estimates, minimum support, and explicit regression constraints.
Future studies should compare these designs under matched rollout and token budgets and report candidate yield, acceptance calibration, family-level effects, and deployment cost.

\section{Conclusion}

Frozen language agents can benefit from useful text policies.
In our experiments, the difficult part is learning those policies from trajectories.
Richer evidence and broader search help intermediate steps, but they do not yield a reliable end-to-end gain.
Agent-level TextGrad therefore needs better rule generation and better rule selection, not only more feedback.

\section*{Limitations}

Our evidence covers two 7B models and three text benchmarks.
It may not transfer to web, software, robotics, or safety-critical agents.
The human rules and their family routing are privileged, so their gain is an upper-bound capacity test rather than an autonomous method.
The structured controls are even narrower because some rule phrases activate pre-written routines.

The trace study keeps every step but shortens some observations and valid-action lists.
The counterfactual study produces only seven valid rules, which limits conclusions about the update operator.
Its validation sets contain eight instances per family.
The GEPA test uses one benchmark, the same 7B model for acting and reflection, and more construction rollouts than the single-rule studies.
We do not provide matched tests of RAPOA, Reflexion, ExpeL, or weight-training methods.
Finally, the intervals are exploratory and do not account for multiple comparisons, tokens, latency, energy, or monetary cost.

\section*{Ethical Considerations}

The environments are synthetic and contain no personal data.
A learned rule can still spread one error across many later tasks.
Practical systems should record where each rule came from, test protected behavior before deployment, restrict what an update may change, and support rollback.

\bibliography{references}

\appendix
\section{Policy, Gate, and Comparator Details}

\paragraph{Curated basis and detector.}
The capacity test uses a small human-written rule library.
Table~\ref{tab:basis} reports its size and coverage.
For real models, each rule is ordinary prompt text.
For structured actors, some phrases turn on matching routines already present in code.

The routing is also human-written.
For each failed training record, a benchmark-specific detector chooses the rule for that task family and adds it once.
The test therefore measures whether an actor can use a known rule under a known family label.
It does not measure autonomous diagnosis or retrieval.

Representative rules describe objective order and graph exploration in TextWorld, arithmetic and map navigation in TextWorldExpress, and game strategies in TextArena.
The supplement contains the exact text in \mbox{\texttt{RULE\_TEXTS}}, \mbox{\texttt{RULE\_BY\_GAME}}, \mbox{\texttt{ENV\_RULES}}, and \mbox{\texttt{SLM\_HINTS}}.

\begin{table}[h!]
\centering
\scriptsize
\setlength{\tabcolsep}{2.5pt}
\begin{tabular}{lrrl}
\toprule
Setting & Basis & Selected & Trigger \\
\midrule
Structured TW & 3 & 3 & failure family \\
Structured TA & 12 & 6 & negative advantage \\
Real TW & 3 & 3/3 & keyed failure \\
Real TWExpress & 8 & 8/8 & fail/invalid/repeat \\
Real TextArena & 12 & 10/12 & fail/invalid/truncate \\
\bottomrule
\end{tabular}
\caption{Curated rule counts. Real-model counts are identical for both actors. ``Selected'' is not generated text: the detector activates a human-authored rule keyed to the benchmark family. Real TextArena also retains its standard environment hint in the fixed condition.}
\label{tab:basis}
\end{table}

\paragraph{Mechanism control: structured actors (not text-only).}
This appendix-only study asks whether storing a known control choice can save later interaction when the interface already implements that choice.
These actors are not ordinary language models: a rule phrase can activate a matching routine already present in code.
We therefore exclude these results from the real-agent tables and central claim.
They demonstrate a possible efficiency mechanism, not an effect for text-only agents.
At matched success, a stored policy uses 46.5\% fewer actions than retry with persistence on TextWorld and 41.0\% fewer on TextArena (Table~\ref{tab:controller-controls}).
A random policy of the same size performs worse on TextArena, so prompt length alone does not explain the gain.

\begin{table}[h!]
\centering
\scriptsize
\setlength{\tabcolsep}{2.5pt}
\begin{tabular}{llrrr}
\toprule
Suite & Method & Success & Attempts & Actions \\
\midrule
TextWorld & Stored policy & 56.7 & 1.00 & \textbf{42.25} \\
TextWorld & Retry + persist & 56.7 & 1.48 & 78.94 \\
\midrule
TextArena & Stored policy & 61.5 & 1.00 & \textbf{10.19} \\
TextArena & Random rules & 42.8 & 1.00 & 11.19 \\
TextArena & Retry + persist & 61.5 & 1.43 & 17.28 \\
\bottomrule
\end{tabular}
\caption{Mechanism control with routine-triggering structured actors: success (\%), attempts, and mean test actions. Random is the mean over 30 policies with six rules each. These results are not directly comparable to the text-only language-agent results.}
\label{tab:controller-controls}
\end{table}

The broader structured suite has the same limit.
Supported interfaces improve sharply, while generic fallback tasks do not change (Table~\ref{tab:secondary-results}).
These controls show policy execution inside known interfaces, not general transfer.

\begin{table}[h!]
\centering
\scriptsize
\setlength{\tabcolsep}{3pt}
\begin{tabular}{lrrr}
\toprule
Suite & Success & Invalid & Turns \\
\midrule
Broad-50 overall & 18.5$\to$36.9 & 51.8$\to$45.1 & 9.32$\to$8.50 \\
\textbf{Supported} & \textbf{16.7$\to$66.7} & \textbf{22.2$\to$4.2} & \textbf{13.29$\to$11.07} \\
Generic fallback & 19.5$\to$19.5 & 69.1$\to$69.1 & 7.00$\to$7.00 \\
\textbf{Difficulty} & \textbf{16.7$\to$83.3} & \textbf{13.9$\to$0.0} & \textbf{22.60$\to$16.96} \\
MiniWoB-50 & 20.0$\to$27.3 & 2.0$\to$2.0 & 4.05$\to$3.86 \\
\bottomrule
\end{tabular}
\caption{Descriptive mechanism controls with structured actors (\%). Bold rows are the largest changes. These rows have no confirmatory intervals and support no claim about text-only agents.}
\label{tab:secondary-results}
\end{table}

Policy construction uses 7,965 TextWorld and 7,921 TextArena training actions.
The estimated break-even points are 202 and 756 deployments.
Only TextWorld reaches its break-even point in this evaluation.

\paragraph{Retry semantics.}
\label{sec:baseline-details}
The retry control first runs the fixed policy.
If that attempt succeeds, it stops.
If it fails, the task receives one more attempt with the matching human guidance.
Reported turns include both attempts, and the second attempt determines the final outcome.
This control does not generate a Reflexion-style critique, retrieve ExpeL-style training memories, or keep lessons discovered during test.

\paragraph{Context-bounded trace protocol.}
\label{sec:trace-ablation}
For each model and each of eight TextWorldExpress families, both learners see the same three training trajectories and generate one rule at temperature 0.
The summary learner sees the final outcome and the last 12 actions.
The trace learner sees every observation, action, and reward, plus at most 20 valid actions per step.
We shorten observations only when the family evidence would exceed 90,000 characters in a 32,768-token context.

The study keeps all 999 steps, executed actions, and rewards.
It shortens 220 observations, or 22.0\% of steps, while retaining 88.1\% of observation characters overall.
It retains 9,932 of 12,042 valid-action candidates.
Each learned rule is frozen and tested on ten paired seeds, giving 160 episodes per method and 16 model-family clusters.
Because the trace is still bounded, this study measures the effect of richer evidence rather than a truly unbounded trace.

\begin{table}[h!]
\centering
\scriptsize
\setlength{\tabcolsep}{2.5pt}
\begin{tabular}{lrrrr}
\toprule
Model & Fixed & Summary & Bounded & Bounded $-$ summary \\
\midrule
Mistral-7B & 16.25 & 13.75 & 16.25 & +2.50 [$-2.50$, 8.75] \\
Qwen-7B & 17.50 & 8.75 & 13.75 & +5.00 [1.25, 10.00] \\
Pooled & 16.88 & 11.25 & 15.00 & +3.75 [0.00, 8.12] \\
\bottomrule
\end{tabular}
\caption{Success (\%) when the learner sees a compact summary or a context-bounded step trace. The final column gives paired 95\% intervals.}
\label{tab:trace-ablation}
\end{table}

\paragraph{Generated-rule taxonomy.}
We manually code the 32 directly comparable rules from the evidence study: two models, eight families, and two evidence formats.
A \emph{grounded strategy} has a detectable trigger, an executable response, and a relation that should persist within the family.
A \emph{generic heuristic} is plausible but redundant or too vague to select the next action.
An \emph{instance shortcut} hard-codes a room, object, direction, operation, or sequence that need not persist.
A \emph{semantic error} encodes the wrong objective or causal relation.
A \emph{malformed} rule has a conflicting condition and action or returns an unconditional script.
We assign one dominant label without using held-out outcome.
The audit is descriptive, single-coder, not blind, and not independently double-coded.

\begin{table}[h!]
\centering
\scriptsize
\setlength{\tabcolsep}{4pt}
\begin{tabular}{lrrr}
\toprule
Primary label & Summary & Trace & Total \\
\midrule
Grounded strategy & 3 & 4 & 7 \\
Generic heuristic & 5 & 4 & 9 \\
Instance shortcut & 3 & 2 & 5 \\
Semantic error & 5 & 4 & 9 \\
Malformed/unexecutable & 0 & 2 & 2 \\
\bottomrule
\end{tabular}
\caption{Manual taxonomy of 32 generated TextWorldExpress rules. Counts are descriptive and mutually exclusive; several rules exhibit secondary problems.}
\label{tab:rule-taxonomy}
\end{table}

Only two rules are malformed, so formatting is not the dominant failure.
Fourteen are semantically wrong or instance-specific, and nine more are insufficiently specific.
The trace condition raises grounded strategies only from 3/16 to 4/16.
It fixes some causal omissions: Qwen's summary rule for Simon Says invents a transition between fruit actions, while its trace rule correctly reacts to the observed ``Simon says'' command.
But richer evidence can still be abstracted badly.
Mistral's trace rule for coin conditions on an open south door and then says to take a coin even when none is visible; its map rule is an unconditional action list.

The other operators show the same failure types.
GEPA's Qwen policy mainly accumulates generic restatements of the seed prompt.
Its Mistral policy introduces unsupported affordances such as asking another agent for help or using a key or crowbar.
The counterfactual audit also separates proposal quality from selection quality: the gate accepts a shortcut-like coin rule that lowers completion by 20 points, yet rejects a grounded arithmetic rule that raises completion by 5 points.
The qualitative evidence therefore points to three distinct problems: semantic wrongness, overfitting to observed details, and advice that is too vague to change an action.

\paragraph{Exploratory per-family outcomes.}
The primary capacity result pools all eight families.
Table~\ref{tab:family-contribution} shows its heterogeneity rather than eight additional claims.
Every episode receives the complete policy library.
The rows therefore report family-level outcomes, not the isolated effect of one rule.
Each interval hierarchically resamples the two models and then the ten paired seeds within each sampled model.
With only 20 pairs and two model clusters per family, the intervals are unadjusted, low-powered, and not confirmatory.

\begin{table}[h!]
\centering
\tiny
\setlength{\tabcolsep}{1.5pt}
\begin{tabular}{lrr}
\toprule
Family & Success $\Delta$ [95\% CI] & Turns $\Delta$ [95\% CI] \\
\midrule
arithmetic & +5 [$-10$, 25] & $-5.00$ [$-8.30$, $-2.50$] \\
coin & +15 [$-15$, 45] & $-5.70$ [$-15.40$, 2.20] \\
cookingworld & 0 [0, 0] & +2.20 [$-6.15$, 9.90] \\
mapreader & +5 [0, 20] & $-2.05$ [$-6.70$, 0.00] \\
peckingorder & 0 [0, 0] & +0.45 [$-3.60$, 3.65] \\
simonsays & +5 [0, 20] & +0.20 [0.00, 0.80] \\
sorting & +10 [0, 25] & $-11.25$ [$-19.30$, $-4.85$] \\
twc & 0 [0, 0] & 0.00 [0.00, 0.00] \\
\bottomrule
\end{tabular}
\caption{Curated policy minus fixed by family: success in percentage points and mean turns. Each row has 20 paired episodes across two real models. Intervals are unadjusted hierarchical paired-bootstrap intervals and are exploratory.}
\label{tab:family-contribution}
\end{table}

\paragraph{Selection protocol and audit.}
\label{sec:gate-diagnostics}
The original gate combines several outcomes into one score.
It rewards return and success and penalizes errors, repetition, and turns.
For text games, the score is $r+0.5s-0.4i-0.2p-0.1q-0.004T$, where $s$ is success, $i$ is an invalid action, $p$ is a parse failure, $q$ is repetition, and $T$ is turns.
TextArena uses $r+0.5s-0.5i-0.25u-0.005T$, where $u$ is truncation.
These formulas were fixed before test but were not calibrated or preregistered.

Each model-benchmark cell tests four rules on 96 TextArena, 32 TextWorld, or 64 TextWorldExpress development episodes.
It keeps the best rule only if its score beats the fixed policy by at least 0.001.
Across six cells, this process proposes 24 rules and deploys three.

The replication uses a simpler rule that was locked before results were known.
On eight development instances per model-family, it first compares success, then normalized return, and finally turns.
It never accepts a rule merely because it reaches the same failed outcome faster.

\begin{table}[h!]
\centering
\scriptsize
\setlength{\tabcolsep}{3pt}
\begin{tabular}{lrrl}
\toprule
Study & Proposed & Accepted & Held-out audit \\
\midrule
Original & 24 & 3 & 0 clear positive cells \\
Replication & 7 & 3 & 1 harmful; 1 helpful rejected \\
\bottomrule
\end{tabular}
\caption{Selection audit. The original gate uses benchmark-level development sets; the replication uses eight instances per family.}
\label{tab:gate-audit}
\end{table}

\begin{table}[h!]
\centering
\scriptsize
\setlength{\tabcolsep}{2.5pt}
\begin{tabular}{llrr}
\toprule
Model & Benchmark & Ungated $\Delta$ & Gated $\Delta$ \\
\midrule
Qwen & TextArena & +0.8 [0.0, 2.5] & 0.0 [0.0, 0.0] \\
Qwen & TWExpress & $-2.5$ [$-6.3$, 0.0] & +1.3 [$-3.8$, 7.5] \\
Qwen & TextWorld & \textbf{+2.5 [0.4, 4.6]} & 0.0 [0.0, 0.0] \\
Mistral & TextArena & $-6.7$ [$-25.0$, 5.0] & 0.0 [0.0, 0.0] \\
Mistral & TWExpress & +1.3 [0.0, 3.8] & +2.5 [0.0, 7.5] \\
Mistral & TextWorld & $-2.1$ [$-4.6$, 0.4] & 0.0 [0.0, 0.0] \\
\bottomrule
\end{tabular}
\caption{Original generated-policy success deltas in percentage points relative to fixed prompting. Ungated always deploys the edit; gated may retain the base.}
\label{tab:generated}
\end{table}

A post-hoc sensitivity test uses saved structured TextWorld trajectories.
Raising the required score gain from 0.001 to 0.05 and 0.20 reduces acceptance from 10/10 to 9/10 and 6/10.
Test success falls from 56.7\% to 54.2\% and 46.2\%.
A gate that protects every family accepts 7/10 rules and reaches 49.2\%.
Stricter gates can therefore reject helpful changes.
Because this analysis reuses development data, it does not establish a better gate.
Paired confidence bounds, family-level regression checks, and abstention are promising, but we do not validate them here.

\paragraph{Counterfactual-credit protocol and outcomes.}
\label{sec:counterfactual-credit}
The counterfactual study uses training seeds 397001--397002, development seeds 407001--407008, and test seeds 417001--417020.
The branch and retry learners share the same actor, rule format, action validator, development gate, and frozen test policy.
Branching finds 19 same-state action comparisons with different outcomes and turns them into seven valid rules.
Full retries find 41 trajectory pairs with different outcomes and produce five valid rules.
The gate accepts three branch rules and one retry rule.

\begin{table}[h!]
\centering
\scriptsize
\setlength{\tabcolsep}{3pt}
\begin{tabular}{lrrr}
\toprule
Policy & Success & Return & Turns \\
\midrule
Fixed & 15.31 & $-0.164$ & 22.01 \\
Retry-only, gated & 13.75 & $-0.179$ & 22.36 \\
Counterfactual, ungated & 14.37 & $-0.155$ & 22.80 \\
Counterfactual, gated & 14.06 & $-0.159$ & 22.87 \\
\bottomrule
\end{tabular}
\caption{Counterfactual replication with 320 paired episodes per policy. Success is percent; return is normalized environment return.}
\label{tab:counterfactual-credit}
\end{table}

\begin{table}[h!]
\centering
\scriptsize
\setlength{\tabcolsep}{2.3pt}
\begin{tabular}{llrrrr}
\toprule
Model & Family & Support & P & A & $\Delta$S / $\Delta$R \\
\midrule
Qwen & arithmetic & 0 & N & N & 0 / .000 \\
Qwen & coin & 0 & N & N & 0 / .000 \\
Qwen & cookingworld & 1 & N & N & 0 / .000 \\
Qwen & mapreader & 3 & N & N & 0 / .000 \\
Qwen & peckingorder & 0 & N & N & 0 / .000 \\
Qwen & simonsays & 0 & N & N & 0 / .000 \\
Qwen & sorting & 0 & N & N & 0 / .000 \\
Qwen & twc & 2 & Y & Y & 0 / +.119 \\
Mistral & arithmetic & 1 & Y & N & +5 / +.100 \\
Mistral & coin & 2 & Y & Y & $-20$ / $-.200$ \\
Mistral & cookingworld & 1 & Y & N & 0 / $-.007$ \\
Mistral & mapreader & 2 & N & N & 0 / .000 \\
Mistral & peckingorder & 1 & Y & N & 0 / .000 \\
Mistral & simonsays & 0 & N & N & 0 / .000 \\
Mistral & sorting & 2 & Y & Y & 0 / +.150 \\
Mistral & twc & 4 & Y & N & 0 / $-.019$ \\
\bottomrule
\end{tabular}
\caption{Audit of every counterfactual rule. Support counts same-state comparisons with different outcomes. P/A mean proposed/accepted. $\Delta$S and $\Delta$R are changes in test success and normalized return.}
\label{tab:counterfactual-rules}
\end{table}

Two accepted rules leave success unchanged but improve partial return: Qwen \texttt{twc} by 0.119 and Mistral \texttt{sorting} by 0.150.
The accepted Mistral \texttt{coin} rule is harmful: success falls from 5/20 to 1/20.
The gate also rejects a Mistral \texttt{arithmetic} rule that raises success from 1/20 to 2/20.
These two errors explain why the learned policy is not ready for deployment.
The 320 test episodes measure policy outcomes well enough for this study, but only seven valid rules support claims about the update process itself.

\paragraph{Contemporary comparators.}
\label{sec:comparators}
Table~\ref{tab:comparators} separates methods that update text from methods that update model weights.
Our direct comparison uses official GEPA 0.1.1 on fresh TextWorldExpress seeds.
GEPA starts from the same fixed global policy.
It repeatedly reflects on traces, mutates the policy, and keeps candidates that perform well on different development instances.
It receives no human rule library or family routing.

For each model, GEPA searches on 24 training cases, selects on 32 development cases, and freezes one policy for 80 paired test cases.
The same 7B model acts and reflects.
The nominal budget is 320 metric calls per model.
The implementation logs 320 calls for Qwen and 322 for Mistral because it checks the budget between iterations.
Together, the two searches use 642 environment rollouts and 62 reflection calls.

\begin{table}[h!]
\centering
\scriptsize
\setlength{\tabcolsep}{2.5pt}
\begin{tabular}{lrrr}
\toprule
Model & Fixed & GEPA & $\Delta$ [95\% CI] \\
\midrule
Qwen-7B & 20.00 & 18.75 & $-1.25$ [$-7.50$, 5.00] \\
Mistral-7B & 17.50 & 21.25 & +3.75 [$-2.50$, 10.00] \\
Pooled & 18.75 & 20.00 & +1.25 [$-3.13$, 5.63] \\
\bottomrule
\end{tabular}
\caption{Official GEPA versus fixed prompting on held-out TextWorldExpress seeds. Values are success percentages with paired model-family intervals.}
\label{tab:gepa}
\end{table}

GEPA explores seven Qwen candidates and six Mistral candidates.
It improves the best development score for both models, but its pooled held-out interval spans zero (Table~\ref{tab:gepa}).
Normalized return rises by 0.035 (CI: $-0.026$ to 0.096), and turns fall by 0.49 (CI: $-1.44$ to 0.43).
Those intervals also span zero.
This result is inconclusive, not evidence that GEPA fails in general.

We do not run matched versions of RAPOA, Reflexion, or ExpeL.
Feedback Descent also remains unmatched because it requires pairwise textual comparisons.
HiPER and HCAPO assign credit to intermediate steps before updating model weights.
LangMARL learns language policies from multi-agent credit.
JERP, multi-module GRPO, and StarPO also change weights.
A fair comparison with those methods would need to match training compute as well as environment episodes.

\paragraph{Deployment scope.}
We do not test semantic retrieval for individual instances.
Generated rules are shared within a model-family cell.
Human rules use hand-written task routing, and the trace study uses one rule per family.
This family routing should not be read as learned retrieval from the current observation.

\begin{table}[h!]
\centering
\scriptsize
\setlength{\tabcolsep}{2.5pt}
\begin{tabular}{lll}
\toprule
Method & Updated artifact & Search/credit \\
\midrule
RulePI & rule text & 4 edits + mean gate \\
GEPA & prompts/components & reflect, Pareto, merge \\
RAPOA & agent prompts & analyzer + evolution \\
Feedback Descent & arbitrary text & pairwise rationales \\
LangMARL & language policies & causal replay + agent credit \\
JERP & rules + weights & retrieval + joint training \\
Multi-module GRPO & module weights & module-level gradients \\
StarPO & model weights & trajectory-level RL \\
HiPER & model weights & hierarchical advantages \\
HCAPO & model weights & hindsight step values \\
\bottomrule
\end{tabular}
\caption{Conceptual comparison. Only GEPA receives a direct fixed-policy comparison here; this is not a common-budget leaderboard.}
\label{tab:comparators}
\end{table}

\section{Reproducibility Details}

The real-agent study uses Qwen/Qwen2.5-7B-Instruct and mistralai/Mistral-7B-Instruct-v0.3.
Actor decoding is deterministic, with an 8,192-token context unless noted below.
TextWorldExpress uses three training seeds, eight development seeds, and ten test seeds per family.
The oracle study uses test seeds that do not overlap with training.
Every policy is frozen before test records are read.

The rule prompt asks for a reusable lesson of at most 80 words.
It forbids hidden state, test answers, complete walkthroughs, and seed-specific shortcuts.
In the original study, $D$ keeps the outcome, error status, turns, final six actions, and terminal failure.
The same 7B model then produces four candidate JSON edits through $U$.

The trace study is the only run with a 32,768-token context.
Its two evidence conditions share actor trajectories, generation budget, rule length, test seeds, and deterministic rule generation.
The counterfactual study uses two training, eight development, and twenty test seeds per family.
It matches 102 branch rollouts with 102 full retries in the same model-family cells.

GEPA uses actor temperature 0 and reflection temperature 0.7 with a 32,768-token context.
It keeps every executed action and reward.
It normalizes whitespace, limits each observation to 600 characters, and limits valid-action text to 120 characters.
No executed-action text is removed.

Bootstrap intervals cluster TextWorldExpress by family, TextArena by environment, and TextWorld by generated suite.
The supplement contains exact prompts and examples of helpful and harmful generated rules.

\end{document}